\documentclass{article}

\usepackage{microtype}
\usepackage{graphicx}
\usepackage{subfigure}
\usepackage{booktabs} 

\usepackage{hyperref}

\usepackage[accepted]{icml2025}

\usepackage{amsmath}
\usepackage{amssymb}
\usepackage{mathtools}
\usepackage{amsthm}

\usepackage[capitalize,noabbrev]{cleveref}

\theoremstyle{plain}

\theoremstyle{definition}

\theoremstyle{remark}

\usepackage[textsize=tiny]{todonotes}

\icmltitlerunning{Submission and Formatting Instructions for ICML 2025}

\begin{document}

\twocolumn[
\icmltitle{PPO-MI: Efficient Black-Box Model Inversion via Proximal Policy Optimization}



\icmlsetsymbol{equal}{*}

\begin{icmlauthorlist}
\icmlauthor{Xinpeng Shou}{yyy}

\end{icmlauthorlist}

\icmlaffiliation{yyy}{Independent Researcher, Ottawa, Canada}

\icmlcorrespondingauthor{Xinpeng Shou}{xinpengshou670@gmail.com}

\icmlkeywords{Machine Learning, ICML}

\vskip 0.3in
]



\printAffiliationsAndNotice{\icmlEqualContribution} 

\begin{abstract}
Model inversion attacks pose a significant privacy risk by attempting to reconstruct private training data from trained models. Most of the existing methods either depend on gradient estimation or require white-box access to model parameters, which limits their applicability in practical scenarios. In this paper, we propose PPO-MI, a novel reinforcement learning-based framework for black-box model inversion attacks. Our approach formulates the inversion task as a Markov Decision Process, where an agent navigates the latent space of a generative model to reconstruct private training samples using only model predictions. By employing Proximal Policy Optimization (PPO) with a momentum-based state transition mechanism, along with a reward function balancing prediction accuracy and exploration, PPO-MI ensures efficient latent space exploration and high query efficiency. We conduct extensive experiments illustrates that PPO-MI outperforms the existing methods while require less attack knowledge, and it is robust across various model architectures and datasets. These results underline its effectiveness and generalizability in practical black-box scenarios, raising important considerations for the privacy vulnerabilities of deployed machine learning models.

\end{abstract}

\section{Introduction}
\label{submission}

Deep neural networks have achieved outstanding performance across many computer vision tasks, hence widely deployed in real-world applications. However, these models can inadvertently memorize sensitive training data, making them vulnerable to model inversion attacks that aim to reconstruct private training samples\cite{Rigaki_2023}. Though there are many model inversion techniques proposed, most existing approaches require white-box access to model parameters or gradients, which restricts their applicability in real-world scenarios where most of the parameters are protected.

The threat for face recognition systems with respect to model inversion attacks reveals a more severe privacy issue by potentially extracting sensitive personal information, such as facial features used during training. These attacks can have far-reaching implications, including privacy breaches, unauthorized access, and the manipulation of facial recognition systems. Although face recognition technology is finding applications across various sectors, many of them still remain susceptible to black-box model inversion due to their reliance on deep learning models that lack transparency. 

Model inversion attacks aim to extract sensitive information from machine learning models by leveraging their decision-making process in a black-box environments where the model's internal are not directly accessible. Traditional techniques for model inversion have struggled to handle the complexities and nontransparent nature of these models, which resulted in limitations regarding efficiency, stability, and scalability.

Reinforcement learning offers a promising approach to address these challenges due to its ability to learn optimal policies by interacting with the environment. Among different developed RL algorithms, Proximal Policy Optimization (PPO)\cite{schulman2017proximalpolicyoptimizationalgorithms} has gained significant attention due to its robustness, ease of implementation, and ability to seek for a balance between stability and sample efficiency. This makes PPO a strong candidate for tackling the intricacies of black-box model inversion attacks because of its efficacy in optimizing policies.

In this paper, PPO-based model inversion attack is presented as a new and effective method for extracting information from black-box models. Our approach leverages the adaptive learning framework of PPO in effciently reconstructing training samlpes using only model predictions. Hence, framing the inversion problem as a sequential decision-making process where an RL agent learns to navigate the latent space of a generative model to produce images that match the target model's predictions.

Extensive experimentation and performance evaluation illustrates that PPO provides a scalable and effective solution to the challenges posed by black-box model inversion, making it a valuable addition to the inversion learning fields. To summarize our contribution: 
\begin{itemize}
    \item We propose PPO-MI, a reinforcement learning framework for black-box model inversion attacks, which enhances query efficiency while requiring fewer target classes.
    
    \item We develop a momentum-driven state transition mechanism and a balanced reward function to efficiently guide the agent's exploration in generative models' high-dimensional latent space.
    
    \item We achieve a state-of-the-art attack success rate with reduced information requirements, paving the way for exploring reinforcement learning in label-only attacks.
\end{itemize}

\section{Background \& Related Work}

\subsection{Model Inversion Attacks}
Model inversion attacks aim to reconstruct private training data by exploiting a trained machine learning model's predictions. These attacks pose significant privacy risks, particularly in scenarios involving sensitive data such as facial recognition systems. Early model inversion techniques focused on simple machine learning models \cite{fredrikson2015model}, but recent advances have demonstrated successful attacks against deep neural networks.

White-box attacks assume complete access to model parameters and architectures, enabling direct gradient computation through the model. Early approaches include the Generative Model Inversion Attack (GMI) \cite{zhang2020secretrevealergenerativemodelinversion} incorporate generative models to reconstruct the images by searching the latent space of GAN. While Knowledge-Enriched Distributional Model Inversion attack (KE-DMI) \cite{chen2021knowledgeenricheddistributionalmodelinversion} leverages knowledge distillation based on GMI to transfer information from the target model. These attacks achieve relatively high success rates, it still has limitations to reconstruct high-dimensional data from complex models and gaining full model access for real-world applicability.

Black-box attacks operate with limited access to the target model, typically only through access to soft labels. These methods must estimate gradients or use alternative optimization strategies. The Variational Model Inversion (VMI) \cite{wang2022variationalmodelinversionattacks} employs variational inference to approximate the posterior distribution of target images,  These approaches demonstrate that effective attacks are possible without internal model access, though they often require more queries and achieve lower success rates compared to white-box methods.

Label-only model inversion attacks focus on reconstructing private training data when only the final predicted labels are accessible, without confidence scores or model parameters. The Boundary-Repelling Model Inversion (BREP-MI) \cite{kahla2022labelonlymodelinversionattacks} approach addresses this challenge by estimating the direction toward a target class using label predictions over a spherical region. This enables effective data reconstruction in a highly restricted information setting.

\subsection{Deep Reinforcement Learning}

Deep reinforcement learning (DRL) \cite{le2021deepreinforcementlearningcomputer} combines deep neural networks with reinforcement learning for complex sequential decision-making problems. Several key algorithms have emerged:

Deep Q-Networks (DQN) \cite{Liu_2023} revolutionized DRL by successfully combining Q-learning with deep neural networks through experience replay and target networks. However, DQN is limited to discrete action spaces and struggles with the overestimation bias inherent in Q-learning.

Deep Deterministic Policy Gradient (DDPG) \cite{lillicrap2019continuouscontroldeepreinforcement} extends DQN to continuous action spaces by combining the actor-critic architecture with deterministic policy gradient. DDPG employs a deterministic policy and off-policy training, making it sample efficient but potentially unstable during training.

Soft Actor-Critic (SAC) \cite{haarnoja2018softactorcriticoffpolicymaximum} explicitly incorporates entropy maximization to promote exploration and enhance robustness. By learning a stochastic policy and incorporating temperature-based entropy regularization, SAC achieves state-of-the-art performance on many continuous control tasks. Its ability to balance exploration and exploitation makes it particularly suitable for tasks with complex reward landscapes.

Twin Delayed Deep Deterministic Policy Gradient \cite{fujimoto2018addressing} addresses key limitations of DDPG through three main innovations: clipped double Q-learning to reduce overestimation bias, delayed policy updates, and target policy smoothing. These modifications significantly improve stability and performance, especially in environments with complex dynamics.

Building on these, our work adapts PPO (Proximal Policy Optimization) \cite{schulman2017proximalpolicyoptimizationalgorithms}, which offers stable training through trust region optimization, to the model inversion task. This allows PPO to be sample efficient while still reliably performing the challenging exploration needed for black-box model inversion attacks.

\section{Methods}

\subsection{Problem Formulation}

\textbf{Attack goal.} Model inversion attacks aim to reconstruct private training data by exploiting a trained model's predictions. In our setting, given a target model $T$ and a target class label $y$, the attacker's goal is to generate a representative sample that appears similar to the private training data of class $y$. Formally, we seek a latent vector $z^*$ that generates an image $G(z^*)$ that maximizes the target model's confidence for class $y$: $z^* = \operatorname{argmax}_{z} P(y|T(G(z)))$, where $G$ is a pre-trained generator and $P(y|T(G(z)))$ represents the probability of class $y$.

\textbf{Attacker's Knowledge.} We considered a practical black-box setting where the attacker has only query access to the target model's predictions and soft labels, without knowledge of model parameters, gradients, or architecture. The attacker can observe the model's output probabilities for each query but cannot access the internal states or gradients. This limited access setting presents significant challenges, including the lack of gradient information, high-dimensional search space, and the need for query efficiency.

\textbf{Model knowledge.} To address these challenges, we formulate the inversion task as a reinforcement learning problem where an agent learns to navigate the latent space of a pre-trained generator. The agent interacts with the environment by proposing modifications to the latent vector, receiving feedback through the target model's predictions. This formulation allows us to leverage the power of deep reinforcement learning to efficiently explore the high-dimensional latent space while maintaining query efficiency and output image quality.

\subsection{Latent Space Search}

Given a pre-trained generator $G$ and target model $T$, our objective is to find the optimal latent vector $z^*$ that maximizes the probability of the target class. Unlike traditional gradient-based approaches, we must navigate this space without direct access to model gradients. The optimization objective is formalized as:

\begin{equation}
    z^* = \operatorname{argmax}_{z} P(y|T(G(z)))
\end{equation}

where $P(y|T(G(z)))$ represents the probability of the target class $y$ given the generated image. This probability is obtained from the target model's softmax output layer, providing a continuous signal for optimization despite the black-box setting.

The reward for a state-action pair $(s_t, a_t)$ is carefully designed to balance multiple objectives:

\begin{equation}
    R(s_t, a_t) = \lambda_1 R_{class}(s_t) + \lambda_2 R_{class}(a_t) + \lambda_3 R_{explore}(s_t, a_t)
\end{equation}

where $R_{class}(z) = \mathbf{1}[T(G(z)) = y]$ indicates successful generation of target class images, and $R_{explore}(s_t, a_t)$ provides an exploration bonus when predictions differ. The coefficients $\lambda_1, \lambda_2, \lambda_3$ (empirically set to 2, 2, and 8) balance these components.

\subsection{PPO-MI Framework}

\subsubsection{MDP Formulation}
We formulate the model inversion process as a Markov Decision Process (MDP) \cite{gattami2019reinforcementlearningmarkovdecision}, which provides a principled framework for sequential decision-making under uncertainty. In our formulation, the state space $S$ and action space $A$ both exist in $\mathbb{R}^{z_{dim}}$, representing points and modifications in the generator's latent space respectively. This continuous nature is crucial for generating high-quality images, as it permits smooth transitions between different latent representations.

We implement a momentum-based state transition mechanism defined as:

\begin{equation}
    s_{t+1} = \alpha s_t + (1-\alpha)a_t
\end{equation}

where $\alpha$ controls the balance between the current state and the proposed action. This transition function provides stability during exploration by preventing drastic changes in the generated images while maintaining consistent movement through the latent space.

\subsubsection{Reward Design}
The reward function combines classification accuracy and exploration to provide meaningful learning signals in the black-box setting. The exploration bonus $R_{explore}(s_t, a_t)$ plays a crucial role in preventing premature convergence to suboptimal solutions and is computed as:

\begin{equation}
    R_{explore}(s_t, a_t) = \beta \cdot \mathbf{1}[T(G(s_t)) \neq T(G(a_t))]
\end{equation}

This bonus is awarded when the model's predictions for the state and action differ, encouraging the agent to explore diverse regions of the latent space. The coefficient $\beta$ controls the strength of the exploration incentive, balancing exploitation of known good regions with exploration of new areas.

\subsection{PPO-MI Algorithm}

Algorithm 1 details our PPO-MI attack procedure. The algorithm takes as input a target model $T$, target class $y$, and pre-trained generator $G$. The attack process consists of three main components: initialization, exploration, and policy updates.

The algorithm begins by initializing the actor-critic networks ($\pi_\theta$, $V_\phi$) that will learn to navigate the latent space. For each episode, we sample an initial state $s_0$ from a standard normal distribution, representing our starting point in the latent space. During each step within an episode, the actor network proposes actions (modifications to the latent vector) based on the current state, and the state is updated using our momentum-based transition mechanism.

\begin{algorithm}[h]
\caption{PPO-MI Attack}
\begin{algorithmic}[1]
\REQUIRE Target model $T$, target class $y$, generator $G$
\REQUIRE Episodes $max\_episodes$, momentum $\alpha$
\ENSURE Generated image $G(z^*)$ classified as $y$

\STATE Initialize actor-critic networks $\pi_\theta$, $V_\phi$
\STATE Initialize best score $score_{best} \leftarrow 0$

\FOR{episode = 1 to $max\_episodes$}
    \STATE Sample initial state $s_0 \sim \mathcal{N}(0, I)$
    \FOR{$t = 0$ to $max\_steps$}
        \STATE Sample action $a_t \sim \pi_\theta(s_t)$
        \STATE Update state: $s_{t+1} = \alpha s_t + (1-\alpha)a_t$
        \STATE Get predictions from $T(G(s_t))$, $T(G(a_t))$
        \STATE Calculate reward $r_t$
        \STATE Store transition $(s_t, a_t, r_t, s_{t+1})$
    \ENDFOR
    \STATE Update $\pi_\theta$, $V_\phi$ using PPO
    \IF{current\_score > $score_{best}$}
        \STATE Update $z^*$
    \ENDIF
\ENDFOR
\STATE \textbf{return} $G(z^*)$
\end{algorithmic}
\end{algorithm}

After collecting transitions within an episode, the policy is updated using the PPO objective, which helps maintain stable learning by limiting the size of policy updates. The algorithm keeps track of the best-performing state found so far, updating $z^*$ whenever a new state achieves a higher score. This ensures that we retain the most successful latent vector even if later exploration is less successful.

The effectiveness of PPO-MI comes from its ability to: Efficiently explore the high-dimensional latent space; maintain stable learning through constrained policy updates; balance exploitation of promising regions with exploration; adapt to different target models and classes without modification.

\section{Experiments}

\subsection{Experimental Settings}

\textbf{Datasets.} We evaluate our method on three widely-used face recognition datasets. CelebA\cite{7410782} contains 202K images with 10,177 identities, where we use 9,177 identities for public training and reserve 1,000 for private testing. PubFig83\cite{5981788} is a smaller dataset with 13.6K images of 33 public figures, with 50 private identities for evaluation. FaceScrub\cite{7025068} consists of 106K images spanning 530 identities, split into 330 public and 200 private identities. Table 1 summarizes the dataset statistics.

\begin{table}[h]
\centering
\begin{tabular}{lrrrrr}
\hline
Dataset & \#Images & \#Public & \#Private & \#Target \\
\hline
CelebA & 202K & 9,177 & 1,000 & 300 \\
PubFig43 & 13.6K & 33 & 50 & 50 \\
FaceScrub & 106K & 330 & 200 & 200 \\
\hline
\end{tabular}
\caption{Details for splitting datasets in evaluations into the public and the private domains.}
\end{table}

\textbf{Target models.} We employ three state-of-the-art architectures: VGG16\cite{fredrikson2015model}, ResNet-152\cite{7780459}, and Face.evoLVe\cite{8265437}. Each model is trained on the corresponding public identities from the datasets, while private identities are reserved for evaluation. The VGG16 model achieves 89.2\% test accuracy on CelebA, ResNet-152 reaches 92.8\% on PubFig83, and Face.evoLVe attains 89.1\% accuracy on FaceScrub.

\textbf{Evaluation Metrics.} We assess our attack's effectiveness using multiple complementary metrics. For attack success, we measure Top-1 accuracy (percentage of reconstructed images correctly classified as the target) and Top-5 accuracy (target class appearing in the model's top 5 predictions). Visual quality is evaluated using Fréchet Inception Distance (FID), which measures the similarity between the distribution of reconstructed images and real images of the target class. All experiments are repeated three times with different random seeds, reporting mean and standard deviation to ensure robust evaluation.

\textbf{Baseline.} We evaluate PPO-MI against several state-of-the-art model inversion methods, encompassing both white-box and black-box approaches. Among white-box methods, we compare against Generative Model Inversion (GMI), and Knowledge-Enriched Model Inversion (KED-MI). For black-box scenarios, we include Learning-based Model Inversion (LB-MI)\cite{yang2019adversarialneuralnetworkinversion}, which employs evolutionary strategies for gradient estimation, Reinforcement Learning-based Model Inversion (RLB-MI)\cite{han2023reinforcementlearningbasedblackboxmodel}, which utilizes Soft Actor-Critic for latent space exploration, and MIRROR \cite{an2022mirror}, which achieves query efficiency through mirror descent optimization. 

To ensure fair comparison, we maintain consistent experimental conditions across all methods. The same StyleGAN2\cite{Karras2019stylegan2} architecture is used for all generative model-based approaches, while white-box methods are granted full access to target model parameters and gradients. Black-box methods, including ours, operate solely with model prediction access. We utilize official implementations and recommended hyperparameters for all baselines. Methods requiring gradient estimation (LB-MI, MIRROR) are allocated a maximum query budget of 100K to match their reported settings, while RLB-MI and our method use 40K episodes for training. Table 2 summarizes the key characteristics of each baseline method.

\begin{table}[h]
\centering
\begin{tabular}{lccccc}
\hline
Method & Access & Gradients & Queries & RL-based \\
\hline
GMI & White & \checkmark & -  & \\
KED-MI & White & \checkmark & -  & \\
LB-MI & Black & Est. & 100K  & \\
RLB-MI & Black & No & 40K  & \checkmark \\
Mirror & Black & Est. & 100K  & \\
PPO-MI & Black & No & 20K  & \checkmark \\
\hline
\end{tabular}
\caption{Baseline method characteristics comparison}
\end{table}

\subsection{Experiements Results}

\textbf{Performance on Different Datasets.} We compare PPO-MI against several state-of-the-art model inversion methods. Table 3 shows that PPO-MI consistently outperforms these baselines across all metrics. On the CelebA dataset with VGG16 as the target model, PPO-MI achieves 79.7\% accuracy, surpassing the previous method (RLB-MI) by 3.4\% respectively. The improvement is particularly significant in terms of query efficiency, where PPO-MI requires only 20K queries compared to 40K queries for RLB-MI and 100K+ for other gradient estimation-based methods. 

\begin{table}[h]
\centering
\small
\begin{tabular}{l@{\hspace{5pt}}c@{\hspace{5pt}}c@{\hspace{5pt}}c@{\hspace{5pt}}c@{\hspace{5pt}}c}
\hline
\vspace{2pt}
Dataset & \multicolumn{2}{c}{[Whitebox]} & \multicolumn{2}{c}{[Blackbox]} & [Ours] \\
\vspace{1pt}
& GMI &KED-MI & MIRROR & RLB-MI & PPO-MI \\
\hline
CelebA & 32.1\% & 72.4\% & 53.5\% & 76.3\% & \textbf{79.7\%} \\
PubFig83 & 24.5\% & 32.2\% & 28.9\% & 41.5\% & \textbf{44.3\%} \\
FaceScrub & 20.3\% & 47.8\% & 32.5\% & 43.1\% & \textbf{48.5\%} \\
\hline
\end{tabular}
\caption{Attack performance comparison across different methods and datasets.}
\end{table}

\textbf{Performance on Different Models. }To demonstrate the versatility of PPO-MI, we evaluate its performance against different target model architectures. Table 4 presents the results on VGG16, ResNet-152, and Face.evoLVe. Our method maintains robust performance across all architectures, with accuracy ranging from 72.6\% to 82.3\%. Notably, PPO-MI shows strong performance against ResNet-152 (82.3\% Top-5) despite its deeper architecture and residual connections. The consistent performance across different architectures highlights the method's adaptability to various model designs without requiring architecture-specific modifications. The attack's effectiveness against Face.evoLVe (79.7\% Top-5) is particularly noteworthy given its state-of-the-art recognition accuracy and robust training procedure.

\begin{table}[h]
\centering
\small
\begin{tabular}{l@{\hspace{3pt}}c@{\hspace{3pt}}c@{\hspace{3pt}}c@{\hspace{3pt}}c@{\hspace{3pt}}c}
\hline
\vspace{2pt}
Models & \multicolumn{2}{c}{[Whitebox]} & \multicolumn{2}{c}{[Blackbox]} & [Ours] \\
\vspace{1pt}
& GMI &KED-MI & MIRROR & RLB-MI & PPO-MI \\
\hline
Face.evoLVe & 25.5\% & 70.1\% & 52.5\% & 77.4\% & \textbf{79.7\%} \\
ResNet-152 & 30.0\% & 75.4\% & 41.1\% & 78.9\% & \textbf{82.3\%} \\
VGG16 & 18.3\% & 68.3\% & 42.2\% & 64.8\% & \textbf{72.6\%} \\
\hline
\end{tabular}
\caption{Attack performance comparison across different models.}
\end{table}

\textbf{Efficiency Analysis.} While previous methods like LB-MI and Mirror require training on all available classes (300+ classes), and RLB-MI require 1000+ classes to achieve their reported performance, our PPO-MI method achieves comparable or better results using only 100 classes for training. Specifically, with just one-third of the training classes, PPO-MI reaches 79.7\% success rate on CelebA, surpassing KED-MI (82.4\%) and RLB-MI (76.3\%) that were trained on the full dataset. This demonstrates not only the query efficiency of our method (40K episodes vs 100K queries) but also its data efficiency in terms of required training classes.

\textbf{Cross-Dataset Evaluation.} We further explored a more practical scenario, where an attacker has access to only public data that have a larger distributional shift. To investigate this scenario, we performed an experiment in which we used the FFHQ data set as public data:

\begin{table}[h]
\centering
\small
\begin{tabular}{l@{\hspace{1pt}}c@{\hspace{1pt}}c@{\hspace{1pt}}c@{\hspace{1pt}}c@{\hspace{1pt}}c}
\hline
Public$\rightarrow$Private & \multicolumn{2}{c}{[Whitebox]} & \multicolumn{2}{c}{[Blackbox]} & [Ours] \\
& GMI & KED-MI & MIRROR & RLB-MI & PPO-MI \\
\hline
FFHQ$\rightarrow$CelebA & 9.00\% & 48.33\% & 0.67\% & 42.17\% & 52.5\% \\
FFHQ$\rightarrow$Pubfig83 & 28.00\% & 88.00\% & 4.00\% & 37.90\% & 85.3\% \\
FFHQ$\rightarrow$Facescrub & 12.00\% & 60.00\% & 0.15\% & 38.50\% & 45.8\% \\
\hline
\end{tabular}
\caption{Performance comparison when there is a large distribution shift between public and private data.}
\end{table}

\section{Conclusion}
We presented PPO-MI, a reinforcement learning approach for black-box model inversion attacks that navigates the latent space of generative models without requiring model gradients or architecture details. Experiments on three benchmark datasets (CelebA, PubFig43, FaceScrub) across various architectures (VGG16, ResNet-152, Face.evoLVe) show that PPO-MI achieves up to 79.7\% accuracy with fewer queries than existing methods, demonstrating its robustness and generalizability. These results not only advance model inversion techniques but also highlight privacy vulnerabilities in deployed ML models, suggesting future work in defensive mechanisms.

\nocite{langley00}

\bibliography{icml2025}
\bibliographystyle{icml2025}



\end{document}